\documentclass[conference]{IEEEtran}
\IEEEoverridecommandlockouts
\usepackage{cite}
\usepackage{amsmath,amssymb,amsfonts}
\usepackage{algorithmic}
\usepackage{graphicx}
\usepackage{textcomp}
\usepackage{multirow}
\usepackage{xcolor}
\usepackage{url}
\usepackage{bbm}
\usepackage{refcount}

\DeclareMathOperator*{\argmax}{argmax}

\def\BibTeX{{\rm B\kern-.05em{\sc i\kern-.025em b}\kern-.08em
    T\kern-.1667em\lower.7ex\hbox{E}\kern-.125emX}}

\usepackage{fancyhdr}
\begin{document}

\title{%Toward using Task Relevant Features to Mitigate Bias
Using Positive Matching Contrastive Loss with Facial Action Units to mitigate bias in Facial Expression Recognition
%%% PUT THIS THANKS BACK 
% \thanks{Funding acknowledgements redacted for anonymous peer review.}
\thanks{This research is supported in part by the National Research Foundation, Singapore under its AI Singapore Programme (AISG Award No: AISG2-RP-2020-016), and by a Singapore Ministry of Education Academic Research Fund Tier 1 grant to DCO.}
}

\author{

\IEEEauthorblockN{Varsha~Suresh\textsuperscript{1}, Desmond~C.~Ong\textsuperscript{2,3}}
\IEEEauthorblockA{
\textit{\textsuperscript{1}Department of Computer Science, National University of Singapore} \\
\textit{\textsuperscript{2}Department of Psychology, The University of Texas at Austin} \\
\textit{\textsuperscript{3}Department of Information Systems and Analytics, National University of Singapore} \\
varshasuresh@u.nus.edu, desmond.ong@utexas.edu}
}

\maketitle
\thispagestyle{fancy}
\begin{abstract}

%Much work in recent years has shown that machine learning models may learn 
Machine learning models automatically learn discriminative features from the data, and are therefore susceptible to learn strongly-correlated biases, such as using protected attributes like gender and race. Most existing bias mitigation approaches aim to explicitly reduce the model's focus on these protected features. In this work, we propose to mitigate bias by explicitly guiding the model's focus towards task-relevant features using domain knowledge, and we hypothesize that this can indirectly reduce the dependence of the model on spurious correlations it learns from the data. We explore bias mitigation in facial expression recognition systems using facial Action Units (AUs) as the task-relevant feature. To this end, we introduce Feature-based Positive Matching Contrastive Loss which learns the distances between the positives of a sample based on the similarity between their corresponding AU embeddings. We compare our approach with representative baselines and show that incorporating task-relevant features via our method can improve model fairness at minimal cost to classification performance.

%Machine learning models automatically learn discriminative features from the data and therefore are susceptible to learn strongly-correlated bias features along with them. Most existing bias mitigation approaches aim to explicitly reduce the model's focus towards these bias features. Instead, in this work we aim to mitigate bias by explicitly guiding the model's focus towards task-relevant features using domain knowledge which we hypothesize can indirectly reduce the dependence of the model on bias features. We explore bias mitigation in facial expression recognition systems using Facial Action Units as the task-relevant feature. To this end, we introduce Feature-based Positive Matching Contrastive Loss which learns the distances between the positive samples of a sample based on the similarity between their corresponding AU embedding. We compare our approach with representative baselines and show that incorporating task-relevant features can improve fairness while maintaining the classification performance.
\end{abstract}

\section{Introduction}

The performance of machine learning models depends on the quality of the dataset that they are trained on; any bias in the dataset will be learnt by the model \cite{mehrabi2021survey,hutchinson201950}. One example is dataset imbalance: if in a Facial Expression Recognition training dataset, there are more \emph{Male} faces with \emph{Angry} labels than \emph{Female} faces, then given this spurious correlation, the model might learn to associate features associated with \emph{Male} faces (which is task-irrelevant), with the output class \emph{Angry}. Such task-irrelevant cues may be shortcuts that the model learns to achieve better performance, but which reduce generalizability to new data \cite{pezeshki2021gradient,hong2021unbiased,nam2020learning}. Moreover, for moral, ethical, and legal reasons, we may want to ensure that the model is unbiased with respect to certain \textbf{protected} attributes, such as race and gender.

%Machine learning (ML) models' performance is highly subjected to relies heavily on the dataset it is trained upon and research has shown that they are susceptible to the biases in the data \cite{mehrabi2021survey,hutchinson201950}. Biases are cues which the models pick up as an easy route shortcut to achieve better performance but are irrelevant to the task \cite{pezeshki2021gradient,hong2021unbiased,nam2020learning}.

Generally, there are two broad classes of approaches to mitigate bias in machine learning models. The first class of approaches require knowledge of the labels of the protected attribute. If one has labels of the protected attribute, one could try to compensate for the statistical distribution of the attribute \cite{wang2020towards}, use the labels in contrastive-based approaches \cite{hong2021unbiased}, or to ``train out" the bias using adversarial-based approaches \cite{zhang2018mitigating,xu2020investigating,madras2018learning,alvi2018turning}.

%There have been differing approaches to mitigate biases in machine learning models and broadly they can be categorised into two based on whether they require supervision or not with respect to bias labels. One set of supervised methods try to make the model aware of the bias categories by using them  \cite{wang2020towards} and by using it to determine positive and negative samples in contrastive-based approaches \cite{hong2021unbiased}. Another set tries to make the model unaware of the bias-labels using adversarial-based approaches \cite{zhang2018mitigating,xu2020investigating,madras2018learning,alvi2018turning}.

The second class of approaches do not require labels of the protected attribute during training (although for evaluating our bias mitigation techniques, we would still require labels). Most methods in this class use a helper model that is trained with the knowledge of the type of bias-causing features (such as texture biases in CNN models), and the information from these helper models is used to debias the main models \cite{hong2021unbiased,bahng2020learning,cadene2019rubi}. Other approaches utilise the learning dynamics of neural networks, which have been shown to learn biased cues faster than task-relevant features \cite{nam2020learning, pezeshki2021gradient}. 
%
%Unlike previous methods, mitigation without supervision is more realistic \footnote{for testing fairness, bias labels are used.}. 
%In this category of approaches, most methods have a helper model that is trained to be biased using the knowledge of the type of bias-causing features (such as texture biases in CNN models) and the information from these biased models is further used to debias the main models \cite{hong2021unbiased,bahng2020learning,cadene2019rubi} . On the other hand, some approaches in the unsupervised front instead utilise the learning dynamics of neural network which has been shown to learn bias cues faster than informative features relevant to the task \cite{nam2020learning, pezeshki2021gradient}. While our work falls in the category of unsupervised approaches, our approach focuses specifically on the task-relevant features rather than trying to eliminating task-irrelevant features because oftentimes we can't be exhaustive of the which features contribute to the bias cues especially in in-the-wild scenarios. 

In this work, we propose a different approach to reducing bias. Instead of explicitly reducing the model's reliance on task-irrelevant features, we instead propose to guide the model to focus on a set of task-relevant features---and in doing so, reducing bias in the model. In particular, in the case of Facial Expression Recognition, we consider Facial Action Units (AUs) \cite{ekman1980facial}, which have been widely studied in Facial Expression Recognition research \cite{pham2019facial,khorrami2015deep,valstar2006fully,yao2021action}. We introduce feature-based Positive Matching Contrastive Loss, which uses extracted AUs to compute similarities between faces with the same emotion label (i.e., among positive samples). We use these similarities in a contrastive loss to weigh the distances between each sample and its positive samples. We compare our work with representative baselines using two datasets (RAF-DB and IASLab), and find that, compared to existing approaches which do and do not use bias labels, our approach performs very well at reducing bias.

%In this work, we look at how facial expression recognition models is can be debiased via explicit guidance to task-relevant features. In particular, we look at Action Units, which are widely used to provide more context and improve performance \cite{pham2019facial,khorrami2015deep,valstar2006fully,yao2021action} in facial expression research, as the task-relevant feature for bias mitigation. To achieve this, we introduce feature-based Positive Matching loss which uses the AUs extracted from faces to compute similarities between faces with the same emotion label (i.e amongst the positive samples). We use these similarities to weigh the distance between each sample and its positives when training the facial expression classifier. We compare our work with representative baselines using two datasets RAF-DB and IASLab and find it improves performance compared to existing approaches and in some approaches which uses bias labels.

\section{Related Work}

\subsection{Bias mitigation for Machine Learning models}

%While some of the datasets in addition to the task labels contain bias label annotations eg:, for gender bias mitigation in facial expression recognition, bias labels are generally female and male \cite{li2017reliable} and task labels are six basic emotions. We split the existing mitigation approaches based on the whether bias labels are required for bias mitigation. 
We review existing approaches based on whether they require explicit labels of the protected attribute (i.e., labels of the potential bias) during training.

\subsubsection{Bias mitigation with labels} 
%This is done in one of two ways. 
One set of approaches makes the model aware of the bias such as by explicitly predicting bias labels in a multi-task setting  \cite{dwork2012fairness,wang2021understanding}, compensating for the distributional statistics of the protected classes \cite{wang2020towards} or by incorporating the bias labels in the loss computation \cite{hong2021unbiased}. Alternatively, adversarial-based approaches explicitly try to make the model ``blind" to the bias labels by using a confusion loss \cite{zhang2018mitigating,alvi2018turning} or gradient reversal techniques \cite{beutel2017data,madras2018learning}. Although most existing bias mitigation approaches use bias labels for mitigation, this may not always be practical as it is difficult to exhaustively list all the factors that may induce bias in real-life conditions. In addition, a majority of existing datasets do not contain bias label annotations, which may limit the utility of these approaches.

\subsubsection{Bias mitigation without labels} 
Recently, there has been a shift in bias mitigation strategies towards debiasing the models without using bias labels. A majority of these approaches train an auxiliary  model %which forms weak labels 
to debias the primary model \cite{hong2021unbiased,bahng2020learning,cadene2019rubi}. The auxiliary model is generally trained to predict the task using biased features. For example, Convolutional Neural Networks(CNNs) are biased towards textures \cite{li2020shape}, and so one approach to debias object recognition models is to train an auxiliary model to use textures. For instance, \cite{hong2021unbiased} used the distance between samples from embedding space of the auxiliary model to weigh the positives samples in the contrastive objective of the primary model such that when two positives are closer in the embedding space, they should be weighed less in the primary model.

%, for e.g., for debiasing object recognition models approaches use the fact that CNNs are biased towards textures \cite{li2020shape} and train the auxiliary model to recognize objects using texture-capturing models \cite{hong2021unbiased,bahng2020learning,cadene2019rubi}. All the above methods differ in the way information from the biased auxiliary model is passed to the main model, for instance, \cite{hong2021unbiased} used the distance between samples from embedding space of the biased auxiliary model to weigh the positives (samples with same task labels) in the contrastive objective of the primary model  where when two positives are closer in the biased embedding space, they should be weighed less in the primary model.

Another set of approaches harnesses the learning dynamics of neural networks to mitigate bias \cite{pezeshki2021gradient,nam2020learning}. 
If the dataset contains strong biases, then the model learns early on in training to differentiate samples based on these easier-to-learn bias features, which hampers the model's ability to learn task-relevant features \cite{nam2020learning,pezeshki2021gradient,parascandolo2021learning,kim2021biaswap}. 
%\dco{I'm not really convinced by this previous sentence. As in, I don't believe it the way we've written it. I think there's probably more caveats in those papers that we are glossing over.} \reply{I have tried re-writing the statement to make it clearer, section 2 (a simple example) in \cite{pezeshki2021gradient}, is the gist of what I want to convey.}
For example, \cite{pezeshki2021gradient} introduced Spectral Decoupling which uses a L2-regularisation penalty term to decouple the features from the learning dynamics of neural networks, which in turn gives the model a chance to learn from all the features. 
\cite{nam2020learning} used the observations from the model learning dynamics to select easy and hard samples for the main model with the help of an auxiliary model.

%Another set of approaches harnesses the learning dynamics of neural networks to mitigate bias, without requiring the knowledge of the bias-feature \cite{pezeshki2021gradient,nam2020learning}. It has been shown that strongly-correlated bias features are easy to learn and gradient learning is biased to these features. Therefore, when training, the model tends to learn them first irrespective of the presence of other more predictive features \cite{nam2020learning,pezeshki2021gradient,parascandolo2021learning,kim2021biaswap}. \cite{pezeshki2021gradient} introduced Spectral Decoupling which uses just a L2 penalty regularisation term to decouple the features from the learning dynamics of neural networks, which in turn gives the model a chance to learn from all the features. Another work, used the observations from the model learning dynamics to select easy and hard samples for the main model with the help of an auxiliary model \cite{nam2020learning}. 

Our work does not require bias labels nor knowledge of the bias features. Instead of explicitly removing the dependence of the model on the task-irrelevant features, we aim to increase the importance of task-relevant features, obtained using domain knowledge. We hypothesize that doing so would in turn diminish the model's dependence on the bias features.

%While our work falls in this last category where we neither require bias labels nor require the knowledge of the bias-feature, instead of explicitly removing the dependence of the model on bias-features we aim to look increase the importance of task-relevant features using domain knowledge. We hypothesize that doing so can implicitly diminish the models' dependence on the unknown/known easy-to-learn bias-features.

% knowledge of kind of bias and therefore have a specific design for the auxiliary model, some approaches try to mitigate bias implicitly by guiding the model towards informative and predictive features rather than trying to deviate the model's focus on specific spurious features. Most works in this category focus on  Another work, used the observations from the model learning dynamics to select easy and hard samples for the main model with the help of an auxiliary model \cite{nam2020learning}. The auxiliary model is parallely trained to focus on easy samples (i.e samples which contain spurious features which is learnt early on in the training) which is used by the main model to focus on the hard samples.

% While all of the above discussed works explicitly aim to reduce the models' dependence on spurious features either by using bias-capturing models or by understanding the learning dynamics of models. On the contrary, in this work we aim to explicitly shift the model's focus to the informative features by using domain knowledge 

\subsection{Bias mitigation strategies in Facial Expression Recognition}

Facial expression recognition have been widely deployed in various commercial settings such as in automated candidate screening 
%companies such as HireVue\cite{hirevue} 
where companies use videos from applicants to filter candidates based on their facial expressions. 
Recently, these algorithms have come under scrutiny for reinforcing biases against certain groups of people, such as people with disabilities \cite{whittaker2019disability}. Similarly, Rhue \cite{rhue2018racial} investigated two commercial expression recognition software, Face++ and Microsoft AI, and found that these two commercial offerings rated Black basketball players (in their professional website pictures) as displaying much more negative emotions than White players. 
%Consequences of using biased automated facial expression systems especially in high-stake applications is alarming and it's imperative that we move towards developing fairer models.

Yet, there are only a few papers that have focused on mitigating bias in the context of facial expression recognition. One example is \cite{xu2020investigating}, who investigated existing bias-mitigation techniques that use bias labels for gender- and race-bias mitigation in facial expression recognition of six basic emotions, and found promising results for adversarial approaches. 

In a recent study, \cite{chen2021understanding} used Facial Action Units to mitigate annotation bias, which occurs when human annotators extend their personal/social biases into the data annotation. In this work, while we are less concerned with annotation bias, we also use Facial Action Units as the task-relevant feature to develop fairer expression recognition models. %, however, we continue to align with the facial expression recognition task where we accept the given ground truth labels as correct.

\subsubsection{Facial Action Units for Facial Expression Recognition}

We use the Facial Action Coding System (FACS) as the task-relevant features for Facial Expression Recognition. 
FACS \cite{ekman1980facial} is an anatomically-based coding system which codes facial muscular movements into a set of Action Units (AU), which gives a description of facial muscle movements. 
%These descriptions can  to obtain the predictive features for guiding the model. 
Action Units have been widely used in the study of facial expression and perception \cite{cohn2007observer, rosenberg2020face, sayette2001psychometric}. 
%In the field of Emotion Science Facial Expressions is a well-studied face behaviour where plenty of research have been conducted to map the occurrence of a certain AU or a combination of AU to emotion \cite{rosenberg2020face,ekman1980facial,sayette2001psychometric}. Research has shown that these mappings hold true not just in case of posed faces but also in the case of spontaneous facial expressions \cite{sayette2001psychometric}. 

\cite{pham2019facial} used a FACS-based deep learning architecture to retrieve images that display similar facial expressions in FER2013, an in-the-wild facial expression dataset. Another detailed investigation by \cite{khorrami2015deep} found that there is activity in CNN-based FER models in the regions of the face corresponding to the locations of relevant Facial Action Units, which suggests a high correlation between the facial expressions and facial action units. 
Thus, there is literature to support the claim that these extracted AUs are meaningful indicators of facial expression, and hence we may be able to use these AU features as task-relevant features to guide the model towards these features, and away from task-irrelevant features.

%\notes{"suppose a neural network is provided with a chess book containing examples of chess games with the best movements indicated by a red arrow. The network can take two approaches: 1) learn how to play chess, or 2) learn just the red arrows. Either of these solutions results in zero training loss on the games in the book while only the former is generalizable to new games. With no external knowledge, the network typically learns the simpler solution}. 

\begin{figure}[t]
    \centering
    \includegraphics[width=\columnwidth]{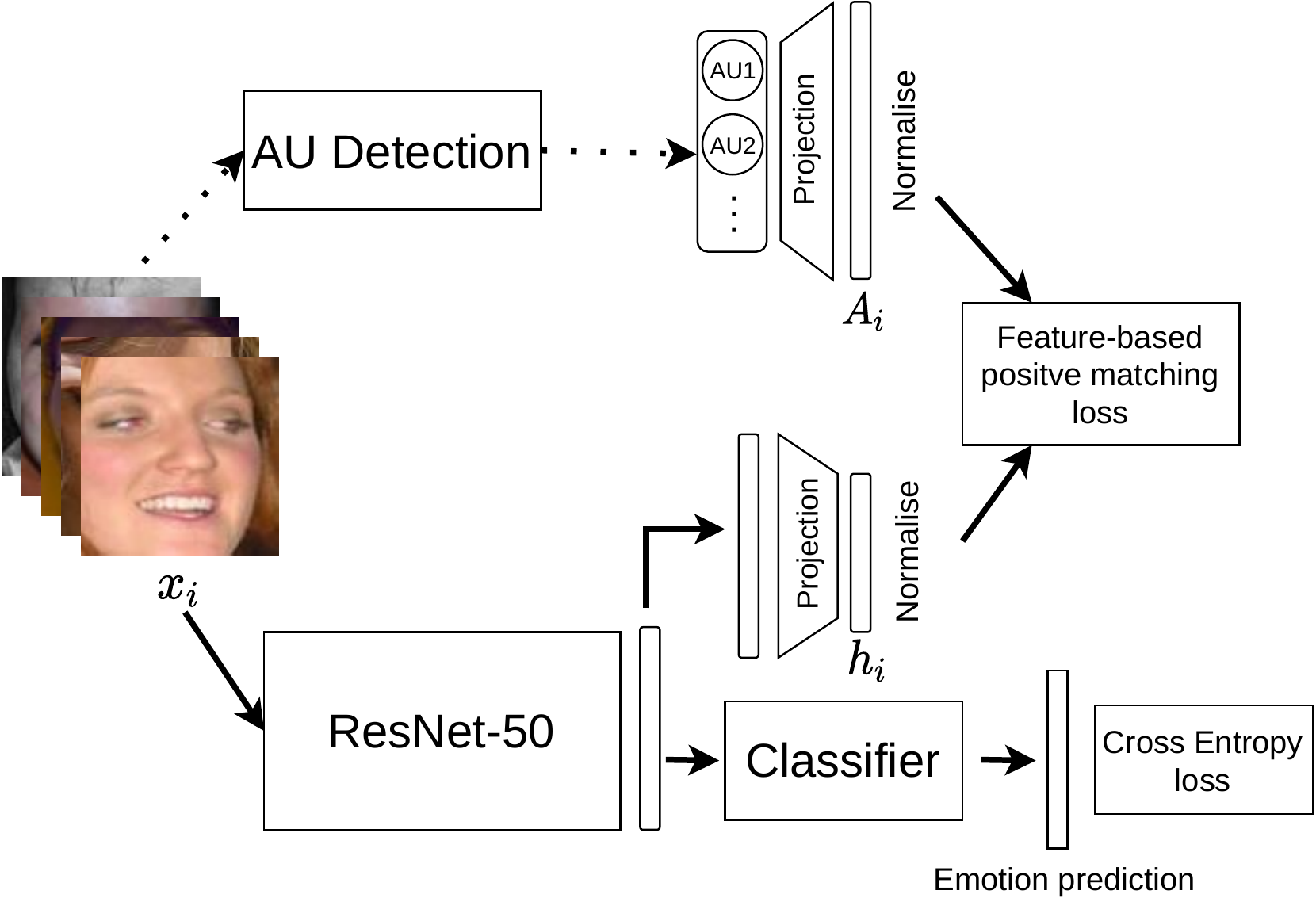}
  \caption{An overview of our proposed Positive Matching Contrastive Loss. $x_{i}$ is the input image, $h_{i}$ is a latent embedding of the input image used to compute the contrastive loss, and $A_{i}$ is AU embedding obtained after projection. Both $h_{i}$ and $A_{i}$ was normalised before computing the loss, in Eqn. \ref{eqn:positive-matching-loss}. Dashed arrows represent \textit{pre-processing} steps (extracting AU from input data) and solid arrows represents connections that are updated during model training. Images shown are taken from the RAF-DB dataset.}
  \label{fig:model} 
\end{figure}

\section{Approach}

\subsection{Model Fairness}
How do we measure if a model is ``fair"? This is not a trivial question, and there have been different notions of fairness discussed in the literature \cite{hardt2016equality,mehrabi2021survey,ghosh2021characterizing}.
In our context, let us consider a model trained to detect happiness (or not), $\hat{y} \in \{0,1\}$, in a dataset containing the sensitive or protected attribute gender, male or female $\{a_{m},a_{f}\} \in A$. We may desire for our model to have the same level of accuracy in predicting ``truly" happy men and as it does for predicting ``truly" happy women, or in other words, to have the same true positive detection rate (of happiness, $y$) across groups $a \in A$. 
%that is, to minimize the discrepancy in the true-positive classifications across all protected groups. 
%to be labelled happy, in other words, just to have equal percentages in all groups within a sensitive attribute we don't want false positives for the minority group. 
%This fairness notion is in line with the existing fairness definition suggested by equality of opportunity,
This is in line with the notion of equality of opportunity \cite{hardt2016equality, mehrabi2021survey}, where we ideally want:
\begin{equation}
    P(\hat{y}=1|y=1,A=a_{f}) = P(\hat{y}=1|y=1,A=a_{m}) \label{Eqn:Equality}
\end{equation}
where $\hat{y}$ is the predicted label, $y$ is the target label, for the protected classes $\{a_f, a_m\} \in A$. In this work, we focus on equality of opportunity as it maintains a balance in achieving fairness by equalising the true positive rates amongst different protected groups \cite{hardt2016equality}. %within a protected attributes while being in line with the supervised learning outcome.

% Demographic parity suggests that same percentages of women and men should be predicted happy irrespective of the target label. This does not ensure fairness especially when there is unequal representation of samples. 
% \begin{equation}
%     P(\hat{y}|A=f) = P(\hat{y}|A=m)
% \end{equation}
% Equality of odds suggests that the given a target label, same percentages of women and men should be predicted happy. This is a more stricter form of fairness as it's difficult to achieve. 
% \begin{equation}
%     P(\hat{y}=1|y=y,A=f) = P(\hat{y}=1|y=y,A=m), y \in \{0,1\}
% \end{equation}

\subsection{Feature-based Positive Matching Loss}

In standard supervised classification, the model learns input features that yields the best classification accuracy---sometimes, some of these output-relevant features may correspond to protected attributes such as race and gender. For example, if the training dataset is imbalanced or has other types of bias in it, such that some protected attributes are more correlated with some output classes (e.g., White faces that tend to be labelled as Happy, compared to Black), then the model would learn to associate these attributes with the output (e.g., see \cite{rhue2018racial} for an example for emotion classification and race).

%In the objective function of standard supervised classification, the model is trained to find invariant features amongst all samples with the same label (positive samples) which yields the best performance metric. Some of these automatically found invariant features be easily learnable cue corresponding to sensitive attributes such as race and gender. For instance, in the case of imbalanced distribution of samples there are more training samples of White faces which is labelled Happy as compared to Black faces. This may cause the model identify being White as one of the invariant features to be labelled happy, guiding the model to make more biased decisions. 

In this work, we propose a different approach. How about if we guide the model to pay attention to certain types of input features that we know, from theory, relate to the output? In the case of emotion classification, there has been systematic research characterizing facial muscle movements, which has been formalized in the Facial Action Coding System \cite{ekman1980facial, cohn2007observer} and provides an informative set of features from which to infer emotions. 

We introduce feature-based Positive Matching Contrastive Loss %(\emph{pos-match}) 
which provides extrinsic guidance to the model using task-relevant features, without the need for explicit bias labels. We hypothesize that such guidance will help the model focus on task-relevant similarities (such as facial muscle movements) over task-irrelevant features (that may be associated with race and/or gender).

%In this work we introduce Feature-based Positive Matching Loss (pos-match) which provides extrinsic guidance to the model using Facial Action Units without the need for demographic labels. We hypothesise that this guidance can help the model focus on demographic independent similarities (such as facial muscle movements) within a given emotion label.

Let the latent embedding of a sample $i$ using a Convolutional Neural Network-based feature extractor be $h_{i}$, and let the embeddings of its positive samples (i.e., the other samples $\{p\} \in P_i$ that share the same output class as $i$) be $\{h_{1}, \cdots, h_{P_i}\}$. We use a similarity function to calculate the pair-wise similarities between $i$ and its positives $\{p\} \in P_i$, to obtain: $\{S(h_{i},h_{1}),\cdots, S(h_{i},h_{P_i})\}$. Next, we repeat the same similarity calculations with the AU embeddings of sample $i$, $\boldsymbol{A}_i$, along with its positives $\{\boldsymbol{A}_1 \cdots, \boldsymbol{A}_{P_i}\}$, and we use these to weight the similarity scores of the $h$ embeddings in the loss:
%
%We use a similarity function $S$, to calculate the similarity vector of between the embeddings to get $\{S(h_{i},h_{1}),\cdots, S(h_{i},h_{P})\}$, where $P$ is the total number of positives for a given sample $x_{i}$ i.e $y_{i} = y_{p}$. Next we weigh the similarity between the positives with respect to a matching score $m_{i,p}$ obtained from the AU embeddings $\boldsymbol{A}$ detected from the input images.
%
\begin{align}
    L_{\text{pos-match}} &= - \frac{1}{N} \sum_{i=1}^{N} \sum_{p=1}^{P_i} S(\boldsymbol{A}_{i},\boldsymbol{A}_{p}) \cdot S(h_{i},h_{p}) \label{eqn:positive-matching-loss}
\end{align}

%\begin{equation}
%    m_{i,p} = S(\boldsymbol{A}_{i},\boldsymbol{A}_{p})
%\end{equation}
%
%\begin{equation}
%    L_{\text{pos-match}} = - \frac{1}{N} \sum_{i=1}^{N} \sum_{p=1}^{P} m_{i,p} \cdot S(h_{i},h_{p})
%\end{equation}

\noindent where $N$ is the total number of samples in a batch. We normalise both AU embeddings $\boldsymbol{A}_{i}$ and latent embeddings $h_{i}$, and we use cosine similarity for $S(.,.)$.

To obtain the AU embeddings, we use the recently-introduced AU detection model, JAA-Net \cite{shao2021jaa}, to get the raw intensities for 12 AUs%
\footnote{AU01: Inner brow raiser, AU02: Outer brow raiser, AU04: Brow lowerer, AU06: Cheek Raiser, AU07: Lid tightener, AU10: Upper Lip Raiser, AU12: Lip Corner Puller, AU14: Dimpler, AU15: Lip Corner Depressor, AU17: Chin Raiser, AU23: Lip Tightener, and AU24: Lip Pressor.} 
in the range of 0 to 1. The AU extraction step is performed prior to the training of the CNN-based feature extractor. During training, each 12-dimensional raw AU vector is projected into a latent space using a non-linear layer with a dimension of 32 with a ReLU %\cite{nair2010rectified} 
activation. This network is trained along with the feature extractor to get the corresponding AU embedding $\boldsymbol{A}_{i}$. 

For feature extraction (Fig. \ref{fig:model}), we fine-tune ResNet50 pre-trained with VGGFace2 weights %except for the classification layer 
\cite{cao2018vggface2,ly2019multimodal,ngo2020facial}. The output of the ResNet-50 network is further projected down to a latent space using a non-linear layer with a dimension of 128 with ReLU activation to obtain $h_{i}$ which is used for computing $L_{\text{pos-match}}$ (Eqn. \ref{eqn:positive-matching-loss}). For facial expression classification, as done in standard fine-tuning, the output from ResNet-50 is passed through a classification layer and trained using a Cross Entropy Loss. Therefore, the combined objective is given by: 
%
%\dco{Varsha: The above paragraph, before my edits, does not correspond to Fig. 1. In Fig 1, the hidden activations after ResNet50 are passed to the classifier, but in the text above it sounds like the $h_{i}$ (used to calculate $L_{\text{pos-match}}$) are passed into the classifier? ----- I have edited it, it's not $h_{i}$ which is used for classification, we follow standard fine-tuning and use the output from the final layer to pass through the classifier.}
%
\begin{equation}
  09  L_{\text{final}} = L_{\text{pos-match}} + L_{\text{CE}} \label{eqn:total-loss}
\end{equation}

\section{Evaluation}
\subsection{Datasets}
We used two emotion-classification datasets with seven labels: \{\emph{happy}, \emph{sad}, \emph{angry}, \emph{fear}, \emph{surprise}, \emph{disgust}, and \emph{neutral}\}. 
\begin{itemize}
    \item \textbf{RAF-DB} \cite{li2017reliable,li2019reliable} consists of crowd-sourced face images from the Internet. All the images are labelled with race, gender and age attributes, and we will consider mitigation of bias by \textbf{race} and \textbf{gender} \footnote{In RAF-DB some samples are labelled "Unsure". Following previous works, for fairness evaluation we do not use these samples.}. The train/validation/test split is 9816 / 2455 / 3068 samples respectively. We used the cropped and aligned version provided by the authors. 
    \item \textbf{IASLab}\footnote{Development of the Interdisciplinary Affective Science Laboratory (IASLab) Face Set was supported by the National Institutes of Health Director’s Pioneer Award (DP1OD003312) to Lisa Feldman Barrett.} is a lab-controlled dataset which comprises of posed expressions that also have \textbf{gender} annotations. In this dataset, all images are modified to have eyes centered in the same location\footnote{\url{https://www.affective-science.org/face-set.shtml}}. We use a train/validation/test split of 1,151 / 144 / 145 samples.
\end{itemize}

\begin{table}[]
\centering
\begin{tabular}{lccc}
\textit{Gender} & Male & Female & Unsure \\ \hline
IASLab & 414 (35.9\%) & 740 (64.1\%) & - \\
RAF-DB & 3,893 (40.2\%) & 5,179 (53.5\%) & 609 (6.3\%) \\ \hline
\hline
\textit{Race} & Caucasian & African-American & Asian \\ \hline
RAF-DB & 7,420 (76.6\%) & 767 (7.9\%) & 1,494 (15.4\%) \\ % \vspace{0.1mm}
\end{tabular}
\caption{Breakdown of data by respect to Gender (Top) and Race (Bottom). Only RAF-DB has race annotations.}
\label{tab:race_gender_dist}
\end{table}

% \begin{table}[]
% \centering
% \begin{tabular}{lccc}
% \hline
%  & \textbf{Male} & \textbf{Female} & \textbf{Unsure} \\ \hline
% \textbf{IASLab} & 414 & 740 & - \\
% \textbf{RAF-DB} & 3893 & 5179 & 609 \\ \hline
% \end{tabular}
% \caption{Data distribution with respect to Gender}
% \label{tab:gender_dist}
% \end{table}

% \begin{table}[]
% \centering
% \begin{tabular}{lccc}
% \hline
%  & \textbf{Caucasian} & \textbf{African-American} & \textbf{Asian} \\ \hline
% \textbf{RAF-DB} & 7420 & 767 & 1494 \\ \hline
% \end{tabular}
% \caption{Data distribution with respect to Race}
% \label{tab:race_dist}
% \end{table}

\subsection{Implementation Details}
\subsubsection{AU Detection}
We use the \texttt{py-feat}\footnote{\url{https://py-feat.org/}} \cite{cheong2021py} package to implement JAA-Net for detecting AUs from training images. We first perform face detection using RetinaFace \cite{deng2020retinaface}. Samples with no faces detected are discarded and if multiple faces are detected then we take the largest face. Each image with a detected face will be passed through JAA-Net to obtain a 12-dimensional AU intensity vector.

\subsubsection{Model Training}
We trained our models using Stochastic Gradient Descent with a momentum of 0.9, a learning rate of 1e-03 and batch size of 32. We trained the models for 30 epochs using a step learning rate scheduler with step size of 10 and decay factor $\gamma$ of 0.5, and did early stopping, choosing the best model based on validation accuracy. We resized the the shorter side of images to 256 pixels and perform center cropping of 224 × 224 pixels to shape the input to the feature-extractor \cite{cao2018vggface2}. The images are normalised using each respective dataset’s mean and standard deviation. To prevent over-fitting, we augmented the datasets by performing horizontal flipping for a randomly-chosen 50\% of the images.

\subsection{Evaluation metric}

\subsubsection{Expression Classification}
For classification performance, we use classification accuracy and weighted-F1 score.

\subsubsection{Fairness measure}
% \notes{is tpr the same, so acc = tp+tn /tp+tn+fp+fn? Come back to this. (tn is not there in multiclass definition)}
We use as our fairness measure the equality of opportunity \cite{hardt2016equality} given in Eqn. \ref{Eqn:Equality}. Following previous research \cite{ghosh2021characterizing,xu2020investigating}, we use the worst-case min-max ratio, which is the ratio of the \emph{minimum} accuracy across all protected groups to the \emph{maximum} accuracy across all protected groups. For example, if gender is the protected group and if the classification accuracy of emotions for males is lower than that for females, then the ratio would be the accuracy for males over the accuracy for females. The closer the min-max ratio is to unity, the smaller the disparity between the true-positive rate between protected groups, and hence the ``more fair'' the model is. % and secondly, true positive rate difference between groups.
Formally, if $a_d$ is the protected group with the highest classification accuracy (for emotions):
\begin{align*}
    a_d = \argmax_{a \in K} {\frac{1}{N_{a}}\sum_{i=1}^{N}\mathbbm{1}_{(\hat{y_{i}}=y_{i}|A=a)}}
\end{align*}
where %$C$ is the number of emotion classes, 
$K$ is the number of group defined by the protected attribute, $N$ is the total number of samples, $N_{a}$ is the number of samples belonging to protected group $a$, and $\mathbbm{1}$ is the indicator function, then this worst-case ratio, the Fairness Score, is given by: % disparity in accuracy scores,
\begin{align}
    %F = \text{min}\bigg\{\frac{\frac{1}{N_{a_{k}}}\sum_{i=1}^{N}\mathbbm{1}_{(\hat{y_{i}}=y_{i}|A=a_{k})}}{\frac{1}{N_{a_{d}}}\sum_{i=1}^{N}\mathbbm{1}_{(\hat{y_{i}}=y_{i}|A=a_{d})}}\bigg\} \forall a_{k} \in K | a_{k} \neq a_{d}
    F = \min_{a_{k} \in K \setminus \{a_{d}\}} \bigg\{\frac{\frac{1}{N_{a_{k}}}\sum_{i=1}^{N}\mathbbm{1}_{(\hat{y_{i}}=y_{i}|A=a_{k})}}{\frac{1}{N_{a_{d}}}\sum_{i=1}^{N}\mathbbm{1}_{(\hat{y_{i}}=y_{i}|A=a_{d})}}\bigg\} \label{eqn:fairness}
\end{align}
    % \item $\text{TPR}_{\text{a1-a2}}$ measures absolute true positive rate difference between two groups $a1$ and $a2$,
% \begin{equation}
%     \text{TPR}_{\text{a1-a2}} = |\frac{\sum_{c=1}^{C}\sum_{i=1}^{N}\mathbbm{1}_{(\hat{y_{i}}=c|A=a_{1},\hat{y_{i}}=c)}}{\sum_{c=1}^{C}\sum_{i=1}^{N}\mathbbm{1}_{(y_{i}=c|A=a_{1},\hat{y_{i}}=c)}} - \frac{\sum_{c=1}^{C}\sum_{i=1}^{N}\mathbbm{1}_{(\hat{y_{i}}=c|A=a_{1},\hat{y_{i}}=c)}}{\sum_{c=1}^{C}\sum_{i=1}^{N}\mathbbm{1}_{(y_{i}=c|A=a_{1},\hat{y_{i}}=c)}}|
% \end{equation} 

\begin{table}
\centering
\resizebox{\linewidth}{!}{
\begin{tabular}{llcc|c}
\multicolumn{5}{c}{\textbf{Fairness by Gender}} \\
                                 & IASLab                           & Male Acc   & Female Acc & Fairness \\ \hline
\multirow{2}{5mm}{w/ labels}   & Domain-aware               & 91.5 (2.3) & 97.3 (1.0) & 94.0 (2.8)                         \\
                                 & Domain-unaware             & 89.4 (2.3) & 94.7 (2.1) & 94.4 (3.4)                        \\ \hline
\multirow{4}{5mm}{w/o labels} & Cross Entropy Baseline              & 90.2 (3.5) & 96.1 (2.8) & 93.8 (2.1)                         \\
                                 & Positive Matching (ours)       & 92.8 (1.0) & 95.3 (1.7) & \textbf{97.1 (2.3)}                         \\
                                 & Spectral Decoupling        & 89.4 (2.3) & 92.4 (2.1) & 95.5 (2.8)                         \\
                                 & Spectral Decoupling + Ours & 88.9 (0.9) & 91.4 (2.7) & 96.4 (2.1)    \vspace{2mm}                     \\  %\hline 
                                   &  RAF-DB                          & Male Acc   & Female Acc & Fairness  \\ \hline
\multirow{2}{5mm}{w/ labels}    & Domain-aware               & 82.1 (0.5) & 85.6 (0.7) & 95.9 (1.0)                         \\
                                  & Domain-unaware             & 84.8 (0.5) & 87.2 (0.3) & 97.3 (0.5)                        \\ \hline
\multirow{4}{5mm}{w/o labels} & Cross Entropy Baseline              & 84.7 (0.7) & 87.4 (0.4) & \textbf{96.9 (1.1)}                         \\
                                  & Positive Matching (ours)       & 83.8 (0.7) & 87.4 (0.5) & 96.0 (1.1)                         \\
                                  & Spectral Decoupling        & 81.8 (0.7) & 85.0 (0.8) & 96.3 (0.4)                         \\
                                  & Spectral Decoupling + Ours & 82.2 (0.8) & 85.1 (0.6) & 96.5 (1.5)                         \\ % \hline
\end{tabular}
}
\vspace{0.1cm}
\caption{Overall gender fairness measure for IASLab (top) and RAF-DB dataset (bottom). Values are averaged over 5 runs with standard deviation in parenthesis. Best scores amongst the approaches without bias labels are given in bold.}
%best scores for each class of approaches (with and without bias labels) in bold. }
\label{tab:gender_overall}
\end{table}

\begin{table*}
\centering
\begin{tabular}{llccccccc}
\multicolumn{9}{c}{\textbf{Fairness by Gender, broken down by Emotion}} \\ 
                                  & IASLab                           & Neutral     & Anger       & Disgust     & Fear        & Happiness   & Sadness    & Surprise     \\
                                  & \# of training samples              & 159         & 155         & 162         & 170         & 170         & 172        & 166          \\
                                  & Female : Male ratio        & 66 : 34     & 64 : 36     & 66 : 34     & 64 : 36     & 64 : 36     & 66 : 34    & 59 : 41      \\ \hline \hline
\multirow{2}{5mm}{w/ labels}    & Domain-aware               & 100.0 (0.0) & 81.1 (9.7)  & 93.8 (4.7)  & 82.6 (11.8) & 100.0 (0.0) & 91.5 (5.4) & 98.6 (2.9)   \\
                                  & Domain-unaware             & 95.4 (6.2)  & 72.1 (19.5) & 87.8 (9.6)  & 83.4 (10.1) & 96.7 (6.7)  & 90.2 (3.3) & 95.7 (5.7)   \\ \hline
\multirow{4}{5mm}{w/o labels} & Cross Entropy Baseline              & \textbf{96.9 (3.8)}  & 77.3 (15.9) & 86.0 (8.0)  & \textbf{95.1 (6.1)}  & \textbf{100.0 (0.0)} & 87.3 (6.3) & \textbf{97.1 (3.5)}   \\
                                  & Positive Matching (ours)       & \textbf{96.9 (3.8)}  & \textbf{87.1 (10.7)} & \textbf{91.8 (3.6)}  & 80.1 (7.0)  & \textbf{100.0 (0.0)} & 91.5 (5.4) & \textbf{97.1 (3.5)}   \\
                                  & Spectral Decoupling        & 90.8 (3.1)  & 80.0 (10.0) & 84.0 (13.6) & 90.3 (10.2) & \textbf{100.0 (0.0)} & 88.9 (7.8) & 84.3 (8.3)   \\
                                  & Spectral Decoupling + Ours & 92.3 (0.0)  & 81.1 (9.7)  & 78.0 (4.0)  & 91.8 (6.2)  & \textbf{100.0 (0.0)} & \textbf{95.1 (4.1)} & 84.3 (11.4)  
                                  \vspace{2mm} \\ 
                                  & RAF-DB                           &     &       &     &         &   &     &     \\
                                  & \# of training samples               & 1979       & 538        & 569        & 208         & 3831       & 1536       & 1020        \\
                                  & Female : Male ratio        & 50 : 46    & 65 : 33    & 58 : 39    & 57 : 41     & 59 : 38    & 53 : 28    & 48 : 41     \\ \hline \hline
\multirow{2}{5mm}{w/ labels}    & Domain-aware               & 96.9 (2.1) & 91.9 (5.5) & 86.6 (6.7) & 79.8 (7.8)  & 96.9 (0.6) & 92.4 (3.2) & 91.7 (2.3)  \\
                                  & Domain-unaware             & 96.4 (0.8) & 92.2 (4.2) & 95.3 (4.0) & 81.0 (10.8) & 97.6 (0.9) & 93.8 (1.7) & 94.3 (3.2)  \\ \hline
\multirow{4}{5mm}{w/o labels} & Cross Entropy Baseline              & \textbf{97.6 (1.7)} & 89.4 (2.4) & 90.6 (4.1) & 79.9 (8.0)  & 96.7 (0.8) & 95.0 (3.8) & \textbf{92.9 (1.5)}  \\
                                  & Positive Matching (ours)       & 97.5 (1.0) & \textbf{92.7 (3.7)} & 90.8 (3.4) & \textbf{82.9 (6.4)}  & \textbf{96.8 (1.2)} & \textbf{95.3 (2.4)} & 91.1 (1.1)  \\
                                  & Spectral Decoupling        & 96.1 (2.8) & 89.4 (3.1) & \textbf{93.9 (2.9)} & 75.8 (8.4)  & 96.5 (1.0) & 93.3 (3.7) & 87.6 (1.1)  \\
                                  & Spectral Decoupling + Ours & 95.2 (2.7) & 85.4 (3.8) & 87.6 (7.5) & 68.5 (13.5) & 96.6 (1.1) & 92.4 (1.7) & 87.6 (1.7)  
                                  \vspace{2mm} \\ 
\end{tabular}
\vspace{0.2cm}
\caption{Gender fairness measure for each emotion class for IASLab (top) and RAF-DB (bottom). Values are averaged over 5 runs with standard deviation in parenthesis. Best scores amongst the approaches without bias labels are given in bold.
Ratio values correspond to percentage of samples belonging to that group in training dataset.}
\label{tab:results_gender_emotion}
\end{table*}

\subsection{Comparison Methods}

We compare our approach with two classes of approaches: mitigation approaches that require labels of the protected attribute like race and gender, and mitigation approaches that do not require these labels. We note that our Positive Matching Contrastive Loss method does not require these labels.

\subsubsection{Mitigation approaches that require bias-labels}

\begin{itemize}
    \item \textbf{Domain-aware}. Instead of a (\# of class)-way classifier, this approach trains a (\# of classes $\times$ \# of groups)-way classifier to make the model aware of the bias labels (``fairness by awareness" \cite{dwork2012fairness,yang2020towards}). We replace the final classification layer of the vanilla model to accommodate for the increased classification categories. In the case of IASLab, this becomes a 14-class classification (7 emotion classes $\times$ 2 gender groups), while in the case of RAF-DB, this becomes a 42-class classification (7 emotion classes $\times$ 2 gender groups\footnote{\label{note1}RAF-DB has an ``unsure" category for gender, but to be in-line with existing approaches, we only consider male and female.} $\times$ 3 races). 
    \item \textbf{Domain-unaware}. We follow \cite{zhang2018mitigating} and use gradient reversal on the domain classification to make the model unlearn the domain features. For IASLab, we classify the gender (2-way classification) and RAF-DB we classify both gender-race combined labels (6-way classification\footnotemark[\getrefnumber{note1}]).
\end{itemize}

\subsubsection{Mitigation approaches that do not require bias-labels}

\begin{itemize}
    \item \textbf{Baseline (Cross Entropy Loss)}. This is the standard baseline, which uses a Cross-Entropy Loss to predict facial expressions $L_{CE}$. This is also our approach without the Positive Matching Loss (Eqn. \ref{eqn:total-loss} without $L_\text{pos-match}$). 
    \item \textbf{Spectral Decoupling}. This method \cite{pezeshki2021gradient} mitigates bias without the need of bias labels. A $L2$ penalty is added to the loss term, and training is the same as the vanilla model. Hyper-parameters in this model are the penalty coefficient, set to $2e$-$05$, and the annealing steps, set based on when the vanilla model overfits (400 steps for IASLab and 3,500 for RAF-DB). In addition, we also implemented Spectral Decoupling with our method (Spectral Decoupling + Positive Matching Loss).
\end{itemize}

\section{Results and Discussion}

%We compared our proposed approach with two categories of representative baselines. Firstly, we perform a direct comparison with approaches which do not use bias labels. Vanilla approach and Spectral Decoupling falls in this category. In addition, as spectral decoupling is a generic approach which alters the learning dynamics of the model, we applied it in conjunction with our approach (Spectral decoupling + Ours). Secondly, we also compare our model with models that utilise bias labels to understand how comparable is our performance with them. To evaluate fairness we use both overall fairness which looks at accuracy disparity between groups across the entire dataset and also, emotion-wise fairness which looks at disparity within each emotion class. 

\begin{table*}
\centering
\begin{tabular}{llccc|c}
\multicolumn{6}{c}{\textbf{Fairness by Race} (RAF-DB)}  \\ 
                                  &                            & Caucasian Acc & African-American Acc & Asian Acc  & \multirow{2}{*}{Fairness}  \\ 
                                  & \# of training samples               & 7420          & 767                  & 1494       &                                    \\ \hline \hline
\multirow{2}{5mm}{w/ labels}    & Domain-aware               & 84.2 (0.4)    & 84.5 (1.2)           & 83.4 (0.9) & 98.0 (0.6)                         \\
                                  & Domain-unaware             & 86.1 (0.3)    & 86.8 (1.0)           & 86.2 (0.7) & 98.6 (1.3)                         \\ \hline
\multirow{4}{5mm}{w/o labels} & Cross Entropy Baseline              & 86.3 (0.3)    & 88.6 (1.0)           & 86.5 (1.1) & 96.9 (0.9)                         \\
                                  & Positive Matching (ours)       & 85.8 (0.2)    & 87.8 (1.0)           & 87.0 (0.8) & 97.6 (0.9)                         \\
                                  & Spectral Decoupling        & 83.8 (0.8)    & 84.7 (1.0)           & 84.4 (1.0) & 97.7 (0.8)                         \\
                                  & Spectral Decoupling + Ours & 84.0 (0.2)    & 84.4 (1.0)           & 84.6 (0.5) & \textbf{98.6 (0.4)} \vspace{0.2cm} \\ 
\end{tabular} 
\begin{tabular}{llccccccc}
                                  &                            & Neutral     & Anger       & Disgust     & Fear        & Happiness   & Sadness    & Surprise     \\
                                  & Ca :  Af : As ratio*         & 72 : 11 : 17 & 86 : 5 : 9  & 77 : 5 : 18 & 84 : 5 : 12 & 76 : 8 : 16 & 75 : 7 : 18 & 84 : 7 : 9   \\ \hline \hline
\multirow{2}{5mm}{w/ labels}    & Domain-aware               & 93.9 (3.2)   & 69.0 (10.6) & 63.0 (9.8)  & 58.9 (13.6) & 97.3 (0.9)   & 89.7 (4.0)  & 80.6 (11.8)  \\
                                  & Domain-unaware             & 94.4 (2.7)   & 69.9 (3.0)  & 66.9 (10.6) & 48.8 (9.9)  & 98.1 (1.5)   & 84.5 (3.3)  & 83.3 (7.1)   \\ \hline
\multirow{4}{5mm}{w/o labels} & Cross Entropy Baseline              & 93.8 (2.6)   & 68.6 (7.2)  & 62.3 (7.6)  & \textbf{49.5 (6.8)}  & 96.5 (1.7)   & 92.8 (4.9)  & 80.4 (3.8)   \\
                                  & Positive Matching (ours)       & \textbf{95.2 (3.2)}   & \textbf{72.8 (8.9)}  & \textbf{69.4 (15.2)} & 47.3 (9.6)  & 96.1 (1.5)   & \textbf{92.8 (3.0)}  & 78.2 (3.9)   \\
                                  & Spectral Decoupling        & 93.5 (2.4)   & 64.0 (10.6) & 60.9 (6.5)  & 36.6 (14.8) & \textbf{96.9 (2.0)}   & 90.0 (3.6)  & 76.5 (7.2)   \\
                                  & Spectral Decoupling + Ours & 90.1 (3.7)   & 65.6 (7.4)  & 64.8 (7.6)  & 41.8 (9.3)  & 95.3 (2.1)   & 90.9 (2.1)  & \textbf{81.3 (7.5)}  \\ \hline
\end{tabular}
\vspace{0.2cm}
\caption{Race fairness measures for the RAF-DB dataset, across all classes (Top) and broken down by individual classes (Bottom). Values are averaged over 5 runs with standard deviation in parenthesis. Best scores amongst the approaches without bias labels are given in bold. *Ca=Caucasian, Af=African-American, As=Asian. (The ratio values for fear is over 100 due to rounding.)} %because the split is actually 83.7: 4.8: 11.5}}
\label{tab:results_race_emotion}
\end{table*}

We first consider how the various methods perform on gender fairness, in Table \ref{tab:gender_overall}. Recall that the Fairness score (Eqn. \ref{eqn:fairness}) is the ratio of the minimum classification accuracy within each group to the maximum accuracy, and values closer to 1 indicate more similar performances across groups. To orient the reader, the baseline model using Cross Entropy Loss (i.e., with no bias mitigation), achieves a 90.2\% accuracy for males and 96.1\% accuracy for females on the IASLab dataset; the fairness score is thus 90.2/96.1 = 93.8\%. Compared to this, our Positive Matching Contrastive Loss method yielded a greater fairness of 97.1\%, and was the best performing of all methods. If we examined the fairness for specific emotion classes (Table \ref{tab:results_gender_emotion}), our method also achieves the best fairness for 5 of the 7 classes, and for sadness, Spectral Decoupling plus our Positive Matching loss achieves the best fairness.

For gender fairness on the RAF-DB dataset, we find that neither our method, nor most of the other methods we tried (except for Domain-unaware), improved overall fairness beyond the Cross Entropy Baseline (96.9\%). When we took a closer look at the performance on RAF-DB for specific emotion classes, in Table \ref{tab:results_gender_emotion}, we find that out of the seven classes, our method yielded the best fairness for four classes (\emph{anger}, \emph{fear}, \emph{happiness} and \emph{sadness}).

If we next consider fairness by race (only in the RAF-DB dataset), in Table \ref{tab:results_race_emotion}, the best performing method that does not use bias labels is the Spectral Decoupling of \cite{pezeshki2021gradient} augmented with our Positive Matching Contrastive Loss, which achieves a Fairness of 98.6\%. When we consider fairness for specific emotion classes, we again find that our approach achieves the best fairness on four of the seven classes, (\emph{neutral}, \emph{anger}, \emph{disgust} and \emph{fear}), and for \emph{surprise}, Spectral Decoupling plus our Positive Matching loss achieves the best fairness.

One point of discussion is that the datasets do not have balanced gender and race ratios. For IASLab, the Female:Male ratio is 64:36 (Table \ref{tab:race_gender_dist}), while for RAF-DB it is a little closer at 57:43 (if we remove those with unknown labels), but still has more Female than Male faces. The racial distribution in the RAF-DB dataset is also heavily imbalanced, with almost 77\% of the faces being Caucasian, 15\% Asian, and only 8\% African-American (and these are just three of many races). While overall the fairness scores seem high, these imbalances do translate to some troubling fairness values when we examine specific classes. For example, for race in RAF-DB and considering the classification of \emph{fear}, the baseline cross entropy method achieves a Fairness Score of 49.5\%, which suggests that the true positive rate for identifying \emph{fear} in one race is half that of identifying \emph{fear} in another. (Our method does not improve fairness for this particular class either). 

%With regard to racial bias, in RAF-DB dataset  the overall disparity in number of samples between groups is much higher than the case of gender where ~77\% of the data comprises of Caucasian faces followed by ~15\% Asian faces  with ~8\% African-American faces. Incorporating the positive matching loss into both the Cross Entropy objective function and the Spectral Decoupling objective function improves the worst-group generalisation score with respect to overall race fairness. When looking at emotion-specific fairness, anger and disgust shows significant improvements in fairness, however, similar to IASLab even though fear has similar representation disparity to anger and disgust there a dip in performance (-2.2) is observed. 

%In overall, we see improvement in fairness especially in difficult cases where there is high disparity in number of samples, having said there are some exceptions to this pattern where we see reduced performance such as in case of fear in IASLab. One possible reason could be the variability in depicting a certain emotion by different individuals --- which can mean that even amongst positives there can be high variations in AU patterns \cite{barrett2019emotional}. We aim to look into this further and possibly make the model robust to variability within the positives in our future work.

Finally, we consider the impact of bias mitigation strategies on classification performance. Intuitively, we might expect that optimizing for two objectives (i.e., a fairness objective in addition to classification accuracy) may result in lower classification performance. We can see in Table \ref{tab:_emotion_classification} that this is indeed the case for RAF-DB, where all the bias mitigation strategies underperformed the cross entropy baseline on classification accuracy. The Domain-Unaware method overall had the smallest drop in accuracy (-0.3\%), and our Positive Matching method had the smallest drop in accuracy among the strategies that do not require labels (-0.4\%). For the IASLab dataset, we observed that, in fact, the Domain-Aware method increased classification accuracy by 1.2\%, and our Positive Matching method had the largest increase in accuracy among the strategies that do not require labels (0.3\%). Thus, especially compared to the other bias mitigation strategies that do not require labels, our proposed method not only increases fairness scores of the model, but also maintains classification performance compared to the standard (un-bias-mitigated) classifier.

% the most important contribution of our method is that it maintains the classification performance when compared to other bias-mitigation strategies,
%This work shows that explicitly incorporating task-relevant features using our domain knowledge into deep-learning models helps improve fairness while maintaining performance of the standard classifier. 

\begin{table}
\centering
%\resizebox{\linewidth}{!}{
\begin{tabular}{llcc|c}
\multicolumn{5}{c}{\textbf{Overall Emotion Classification Performance}}                                                     \\
\multicolumn{2}{c}{IASLab}                                     & Acc        & F1 & $\Delta _{CE} $  \\ \hline
\multirow{2}{5mm}{w/ labels}    & Domain-aware               & 95.4 (0.9) & 95.4 (1.0) & +1.2                    \\
                                  & Domain-unaware             & 93.0 (1.5) & 93.0 (1.6) & -1.2                   \\ \hline
\multirow{4}{5mm}{w/o labels} & Cross Entropy Baseline     & 94.2 (2.8) & 94.2 (2.9) & -                      \\
                                  & Positive Matching (ours)   & 94.5 (0.9) & 94.5 (0.9) & \textbf{+0.3}                    \\
                                  & Spectral Decoupling        & 91.4 (1.2) & 91.4 (1.2) & -2.8                   \\
                                  & Spectral Decoupling + Ours & 90.6 (1.8) & 90.6 (1.8) & -3.6                   \\ \hline
\\
\multicolumn{2}{c}{RAF-DB}                                     & Acc        & F1         & $\Delta_{CE} $  \\ \hline
\multirow{2}{5mm}{w/ labels}    & Domain-aware               & 84.1 (0.5) & 83.9 (0.5) & -2.4                       \\
                                  & Domain-unaware             & 86.2 (0.3) & 85.9 (0.3) & -0.3                       \\ \hline
\multirow{4}{5mm}{w/o labels} & Cross Entropy Baseline     & 86.5 (0.3) & 86.3 (0.3) & -                          \\
                                  & Positive Matching (ours)   & 86.1 (0.3) & 86.1 (0.2) & \textbf{-0.4}                       \\
                                  & Spectral Decoupling        & 84.0 (0.7) & 83.8 (0.6) & -2.5                       \\
                                  & Spectral Decoupling + Ours & 84.1 (0.2) & 83.9 (0.2) & -2.4                       \\ \hline
\end{tabular}%}
\vspace{0.2cm}
\caption{Overall emotion classification performance for IASLab (top) and RAF-DB (bottom) using accuracy and weighted F1 score. Values are averaged over 5 runs with standard deviation in parenthesis. Best scores amongst the approaches without bias labels are given in bold. $\Delta _{CE}$ denotes the accuracy difference of a given method with the Cross Entropy Baseline, which has no bias mitigation.}
\label{tab:_emotion_classification}
\end{table}

\section{Limitations and Future Work}
Our main premise is that a valid approach to bias mitigation is to guide the model to focus on a set of task-relevant features instead of protected attributes. In the specific case of facial expression recognition, we can appeal to many decades of work in psychology to suggest that there may be some desirable task-relevant features such as facial Action Units. However, there is evidence that socio-cultural variations exist within the expressions associated with similar emotions. For instance, \cite{fan2021demographic} found significant differences in AU06 and AU12 intensities across gender, race, and age for happy emotion. Similarly, there is evidence that people across cultures regulate their display of emotions differently leading to variations in emotion intensities displayed \cite{matsumoto2009cross,elfenbein2002universality}. In the future, understanding the effects of these differences can help further reduce irrelevant intra-class variations leading to fairer models.

% that are orthogonal to protected attributes: facial Action Units, which are independent of race and gender. \todo{the assumption that the link between AUs and emotions is invariant across cultures and gender is problematic. Socio-cultural display rules (which differ for men and women) and emotion regulation mechanisms should at least be acknowledged. Even the basic AUs can be more or less visible depending on facial morphology.} However, 

%This work is highly dependant on the evidence that there is high correlation between facial Action Units and facial expressions. 
Also, generalising this approach to other domains would require tailored, specific domain knowledge, which may have to be hand-crafted, or perhaps extracted from external knowledge bases. There is plenty of evidence suggesting that incorporating domain-relevant context into deep-learning models leads to better performance \cite{cui2020knowledge,chen2019facial,suresh2021knowledge,zhong2019knowledge,roy2020incorporating}, and in this work we show that using domain knowledge could also help build \emph{fairer} models. 

Future work could more deeply study the quality of the task-relevant features, for example by varying the type/number of the AUs used, or including other external knowledge sources. 
%, we aim to look at varying the types/number of AUs and also other external knowledge sources. 
While we analyzed the performance of our method on specific emotion classes, another important future direction is to understand how we can extend this approach to optimize for fairness in specific classes, which is especially important in cases where we have larger number of classes that may be more closely confusable, such as in fine-grained emotion classification 
%Another important future direction is to understand how we can extend positive matching loss to fine-grained emotion classes which is a more challenging task than basic emotion set explored in the current work
\cite{panda2018contemplating,suresh2021not,demszky2020goemotions,liang2020fine}.

\section{Conclusion}
In this work, we introduced feature-based Positive Matching Contrastive loss which reduces (equality of opportunity) bias in facial expression recognition models by explicitly guiding the model towards task-relevant features---Facial Action Units. Our method aims to bring the hidden representations of positive samples (samples with same emotion label) closer together according to their similarity in task-relevant features (Action Units). Our approach was able to improve fairness at minimal cost to classification performance when compared to existing bias mitigation methods in two commonly used facial expression recognition datasets. This work is a step towards developing fairer machine learning models which is in turn important for the ethical deployment of these models in society. 
%can in turn help the society utilise the benefits provided by these models with reduced harmful consequences.  

\section*{Ethical Statement}

Machine learning models have been widely deployed in multiple settings that directly affect peoples' lives, and has been shown to be biased by race for emotion recognition \cite{rhue2018racial}. There is a pressing need to study bias mitigation strategies, especially in facial expression recognition. Our work is a step in this direction, by offering a way to improve fairness without the need to use labels of the protected attributes (such as race and gender labels), by guiding the model to focus on features that are directly relevant to the task at hand.
%, for eg, in case of law enforcement, where they have been used for surveillance and research has shown that these systems are highly biased with larger number of false positives for Black people \cite{garvie2016perpetual}. This example and many others \cite{hirevue,rhue2018racial,buolamwini2018gender} poses a strong need to build fairer systems especially in situations where the stakes are high such as the above example \cite{buolamwini2018gender}. 
%We consider our work as a step towards making this possible where we show that incorporating domain knowledge can help guide the model towards the right features which can lead to fairer models while maintaining the performance in their primary task. 

%Consequences of using biased automated facial expression systems especially in high-stake applications is alarming and it's imperative that we move towards developing fairer models.

%\section*{Acknowledgment}
\bibliographystyle{IEEEtran}
\bibliography{biblio}

\end{document}